\documentclass[sigconf, cameraready]{acmart}

\usepackage[utf8]{inputenc} 
\usepackage[T1]{fontenc} 
\usepackage{amsfonts}       
\usepackage{microtype}      
\usepackage{tabularx}
\usepackage{multirow}
\usepackage{amsmath}
\usepackage{tabularx}
\usepackage{colortbl}

\usepackage{caption}
\usepackage[list=true]{subcaption}
\usepackage{pifont}
\usepackage{bm}
\newlength{\myMheight}
\usepackage{mathtools}

\newcommand{\sref}[2]{\hyperref[#2]{#1}~\ref{#2}}
\newcommand{\norm}[1]{\left\lVert#1\right\rVert}
\newcommand{\xmark}{\text{\ding{55}}}

\newcommand{\bt}{\fontseries{b}\selectfont}

\newcommand{\mhyphen}[0]{\text{-}}
\DeclarePairedDelimiter\abs{\lvert}{\rvert}%
\AtBeginDocument{%
  }

\copyrightyear{2022}
\acmYear{2022}
\setcopyright{acmlicensed}
\acmConference[AIMLSystems 2022]{The Second
International Conference on AI-ML Systems}{October 12--15, 2022}{Bangalore,
India}
\acmBooktitle{The Second International Conference on AI-ML Systems
(AIMLSystems 2022), October 12--15, 2022, Bangalore, India}
\acmPrice{15.00}
\acmDOI{10.1145/3564121.3564127}
\acmISBN{978-1-4503-9847-3/22/10}

\begin{document}

\title{Efficient Graph based Recommender System with Weighted Averaging of Messages}

\author{Faizan Ahemad}
\email{fahemad3@gmail.com}
\affiliation{%
  \institution{Amazon India}
  \streetaddress{Brigade Gateway, WTC, 8th floor, 26/1, Dr. Rajkumar Road, Malleshwaram(W)}
  \city{Bengaluru}
  \state{Karnataka}
  \country{India}
  \postcode{560055}
}

\renewcommand{\shortauthors}{Ahemad et al.}

\begin{abstract}
  We showcase a novel solution to a recommendation system problem where we face a perpetual soft item cold start issue. 
  Our system aims to recommend demanded products to prospective sellers for listing in Amazon stores. These products always have only few interactions thereby giving rise to a perpetual soft item cold start situation.  Modern collaborative filtering methods solve cold start using content attributes and exploit the existing implicit signals from warm start items. This approach fails in our use-case since our entire item set faces cold start issue always.
  Our Product Graph has over 500 Million nodes and over 5 Billion edges which makes training and inference using modern graph algorithms very compute intensive. 
  
  To overcome these challenges we propose a system which reduces the dataset size and employs an improved modelling technique to reduce storage and compute without loss in performance.
 Particularly, we reduce our graph size using a filtering technique and then exploit this reduced product graph using \textbf{W}eighted \textbf{A}veraging of \textbf{M}essages over \textbf{L}ayers (\textbf{WAML}) algorithm. \textbf{WAML} simplifies training on large graphs and improves over previous methods by reducing compute time to $\frac{1}{7}$ of LightGCN \citep{lightgcn} and $ \frac{1}{26}$ of Graph Attention Network (GAT) \citep{gat} and increasing recall$@100$ by \bm{${\color{black} 66\%}$} over LightGCN and \bm{${\color{black} 2.3\text{x}}$} over GAT. 
\end{abstract}




\maketitle

\section{Introduction}
\looseness-1 Modern recommenders use collaborative filtering (CF) \citep{cf_intro}, which view user-product interaction data as a partially observed matrix and then try to predict the unseen observations using algorithms such as Matrix Factorization \citep{mf}, NMF \citep{nmf}, DeepFM \citep{deepfm}, Neural CF \citep{ncf} and Graph Convolution Matrix Completion (GCMC) \citep{gcmc}. These methods perform well for warm start products which already have many interactions, but they fail to recommend new products with no interactions. Various neural network methods, such as DropoutNet \citep{dropoutnet} and CLCRec \citep{CLCRec}, are used to inject content-based user and product attributes into the CF algorithms to solve cold start issues. 

Graph based CF methods \citep{graph_recsys} view matrix completion as a link prediction problem on graphs like GCMC \citep{gcmc}, GraphSAGE \citep{graphsage}, PinSAGE \citep{pinsage}, Alibaba's recommender system \citep{alibaba_rec}, Graph Attention Network (GAT) \citep{gat} and LightGCN \citep{lightgcn}. All graph methods \citep{gnnrecsurvey} not only look at the current user and product but also their neighbouring users and products over multiple hops on the product graph. Further, the product graph can be constructed using a variety of nodes and edges apart from the user and product nodes. For example, social network data can be incorporated by linking users who are friends or followers in the social network \citep{gnn_for_social}. User-product interactions can be predicted as link prediction task between user and product nodes. 

In our usecase, we faced the challenge of \textbf{perpetual soft item cold start} where every product from candidate set of recommendations have few previous interactions, see \sref{Table}{table:interactions}, but not zero interactions, once the product obtains interactions above a small set threshold it is removed from candidate set of recommendations. CF systems optimize performance on warm start items and solve cold start by using content attributes and links with other warm start items, but we don't have any warm start items. Cold start methods only use content attributes and fail to exploit the graph structure in our product data. Neither of these methods perform well in our case where only few interactions are present for all candidate items.
Various features required by our usecase are listed in \sref{Table}{table:comparison} along with details on which recommendation models support these features.

\enlargethispage{9pt}

We propose improvements over Collaborative Filtering (CF), cold start methods and Graph methods for our domain of \textbf{perpetual soft item cold start} and improve algorithmic performance as well as reduce compute and data storage requirements by making the below contributions:\vspace*{-5pt}
\begin{enumerate}
		\item \textbf{W}eighted \textbf{A}veraging of \textbf{M}essages over \textbf{L}ayers (\textbf{WAML}), refer \sref{Section}{section:method}, obtains a middle ground between warm start and cold start on our perpetual soft item cold start system. WAML operates over our product graph from \sref{Section}{section:pg} which captures relationships in product catalogue and user-product interactions (clicks, views etc.). For cold start, WAML ensures that product content embeddings aren't eclipsed by collaborative signals unlike other CF systems. We infuse product data into WAML embeddings through BERT \citep{bert} embeddings of product title and description.
		\item \textbf{Reduce product graph size} as described in \sref{Section}{section:pg} and \sref{Figure}{figure1}, using an innovative graph reduction mechanism from billions of edges and nodes to $250x$ reduction in edges and 66x reduction in nodes, providing 45x reduction in overall compute while retaining $79.5\%$ of maximum recall, see last $2$ rows of \sref{Table}{table:main_results}.	
		\item WAML doesn't use complex neural networks for graph neighbourhood feature aggregation, see \sref{Figure}{figure2a}, since feature transformations and non-linear activation on each graph neural network layer has no positive effect on collaborative filtering \citep{lightgcn}. It only performs \textbf{layer-wise weighted combination} of node and aggregated neighbourhood features.  
		
		\item Inspired by Semi-supervised learning \citep{ssl_item, simclrv2} and contrastive learning for recommendations \citep{CLCRec} we train WAML embeddings using using a \textbf{graph contrastive loss} over the link prediction task similar. This enables WAML to converge faster while producing K-Nearest Neighbour friendly user and product vectors.
	\end{enumerate}

\begin{table}
	\small
	\caption{Count of products in our candidate dataset $\mathcal{P_\text{r}}$ and how many interactions each product has.}
	\label{table:interactions}
	\centering
	\begin{tabular}{lrrr}
		\toprule
		
			Product Type     & No. of Interactions        & $\%$ of Products \\
		\midrule
				Cold Start & 0  & 14.2$\%$ \\
				Soft Cold Start & $\leq$3  & 47.9$\%$ \\
				Mildly Warmed & $3 \mhyphen 10$   & 37.8$\%$ \\
				Fully Warmed &  $\geq$10  & 0.00$\%$ \\
		\midrule
			All Products & $\sim$2.8 & 100.$\%$ \\
		\bottomrule
	\end{tabular}
\vspace*{-11pt}
\end{table}
\begin{table*}
	\small
	\caption{Features required by our usecase and which methods support them.}
	\label{table:comparison}
	
	\centering
	\begin{tabular}{lccccccc}
		\toprule
		
						     &     NCF  & MF  & GCMC & Content & GAT + & LightGCN & WAML \\
						     &  &			(SVD) & or GAT & based & DropoutNet & & (ours) \\
			\bt Feature		     &  & & & & & & \\
		\midrule
			Cold Start &                        \xmark &            \xmark &    \xmark &        \checkmark &    \checkmark &    \xmark      & \checkmark \\
			Soft Cold Start &                   \xmark &            \xmark &    \xmark &        \xmark &        \checkmark &    \xmark      & \checkmark \\
			Use product attributes &            \checkmark &        \xmark &    \checkmark &    \checkmark &    \checkmark &    \xmark      & \checkmark \\
			Exploit Graph Structure &           \xmark &            \xmark &    \checkmark &    \xmark &        \checkmark &    \checkmark  & \checkmark \\
			Heterogenous Graph &           		\xmark &            \xmark &    \checkmark 	&   \xmark &        \checkmark &    \xmark  & \checkmark \\
			Simple Implementation &             \checkmark &        \checkmark &\xmark &        \checkmark &    \xmark &        \checkmark  & \checkmark \\
			\midrule
			Trains Fast &                       \checkmark &        \checkmark &\xmark &        \checkmark &    \xmark &        \checkmark  & \checkmark \\
			Fast Inference with KNN &           \xmark &            \checkmark &\xmark &        \checkmark &    \xmark &        \checkmark  & \checkmark \\
			Scaling with Edge count &           \checkmark &        \checkmark &\xmark &        \checkmark &    \xmark &        \checkmark  & \checkmark \\
			Scaling with Node count &           \checkmark &        \checkmark &\xmark &	    \checkmark &    \xmark 		&   \xmark  	& \checkmark \\
			Scaling with Node degree &          \checkmark &        \checkmark &\xmark &        \checkmark &    \xmark &        \checkmark  & \checkmark \\
			Low Memory usage		&               \checkmark &        \checkmark &\xmark &        \checkmark &    \xmark &        \xmark		& \checkmark \\
			Low Overall Compute &               \xmark &            \checkmark &\xmark &        \checkmark &    \xmark &        \checkmark  & \checkmark \\
					
		\bottomrule
	\end{tabular}
\end{table*}

\enlargethispage{11pt}
	
\vspace*{-6pt}
\section{Method}
\label{section:method}

\subsection{Building the Product Graph}
\label{section:pg}

\begin{figure*}[!t]
     \centering
     \begin{subfigure}[b]{0.49\textwidth}
         \centering
         \includegraphics[width=0.99\textwidth]{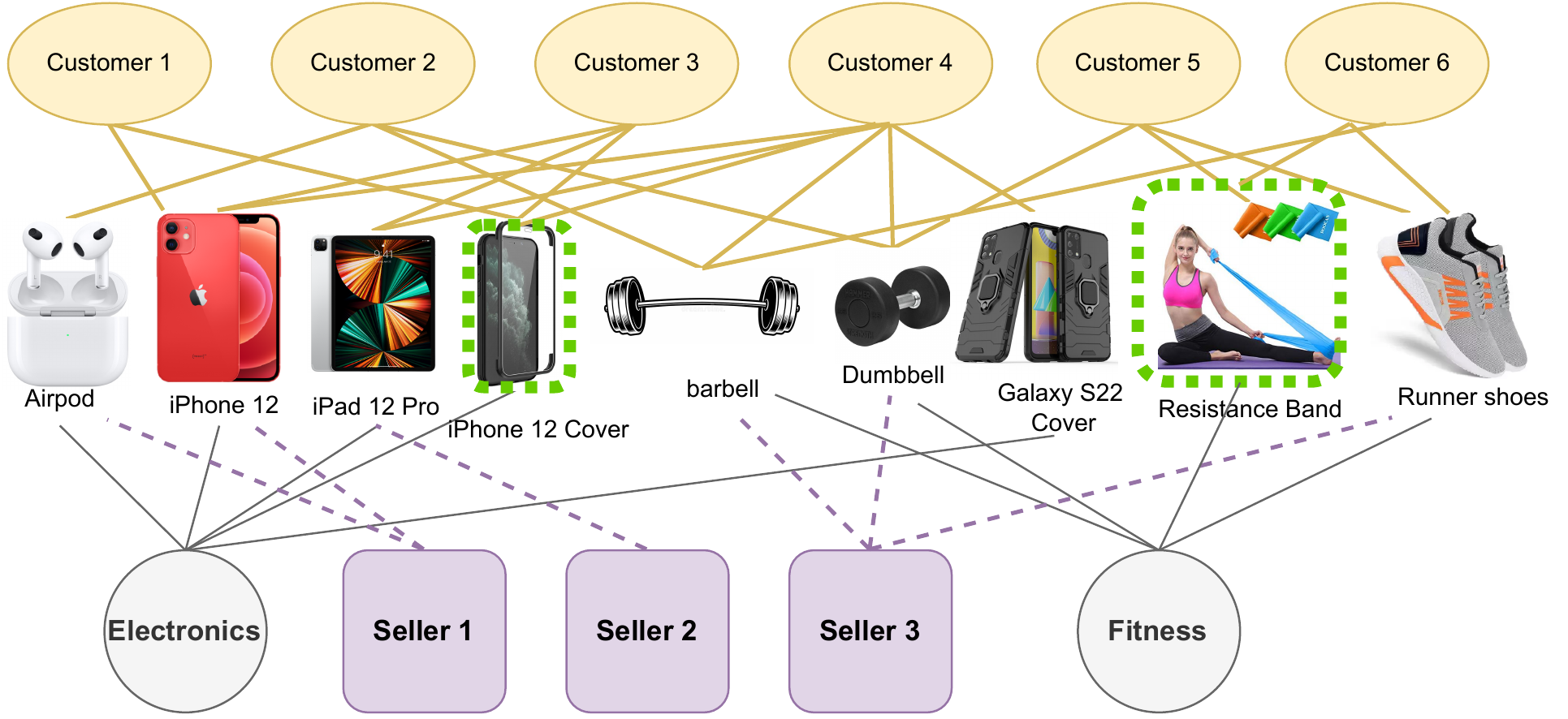}
         \caption{Initial Graph}
         \label{figure1a}
     \end{subfigure}
     \hfill
     \begin{subfigure}[b]{0.5\textwidth}
         \centering
         \includegraphics[width=0.99\textwidth]{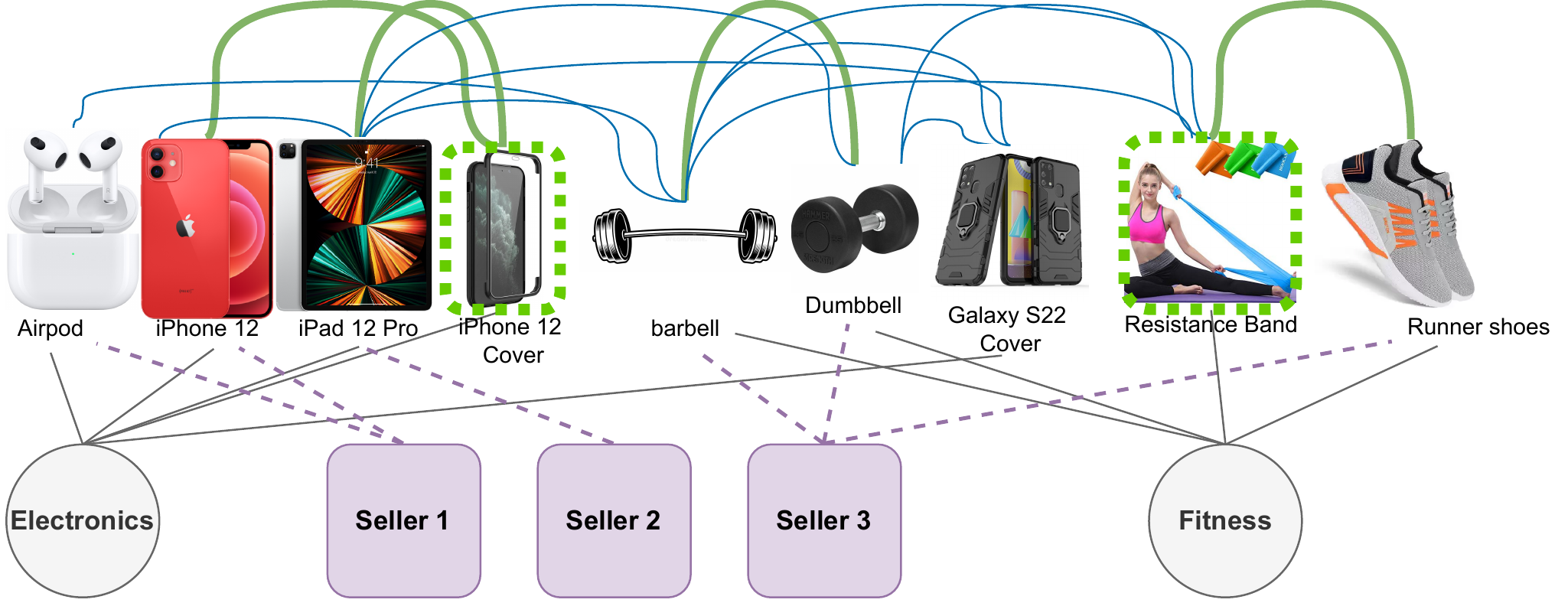}
         \caption{Making product-product edges}
         \label{figure1b}
     \end{subfigure}
     \begin{subfigure}[b]{0.5\textwidth}
         \centering
         \includegraphics[width=0.99\textwidth]{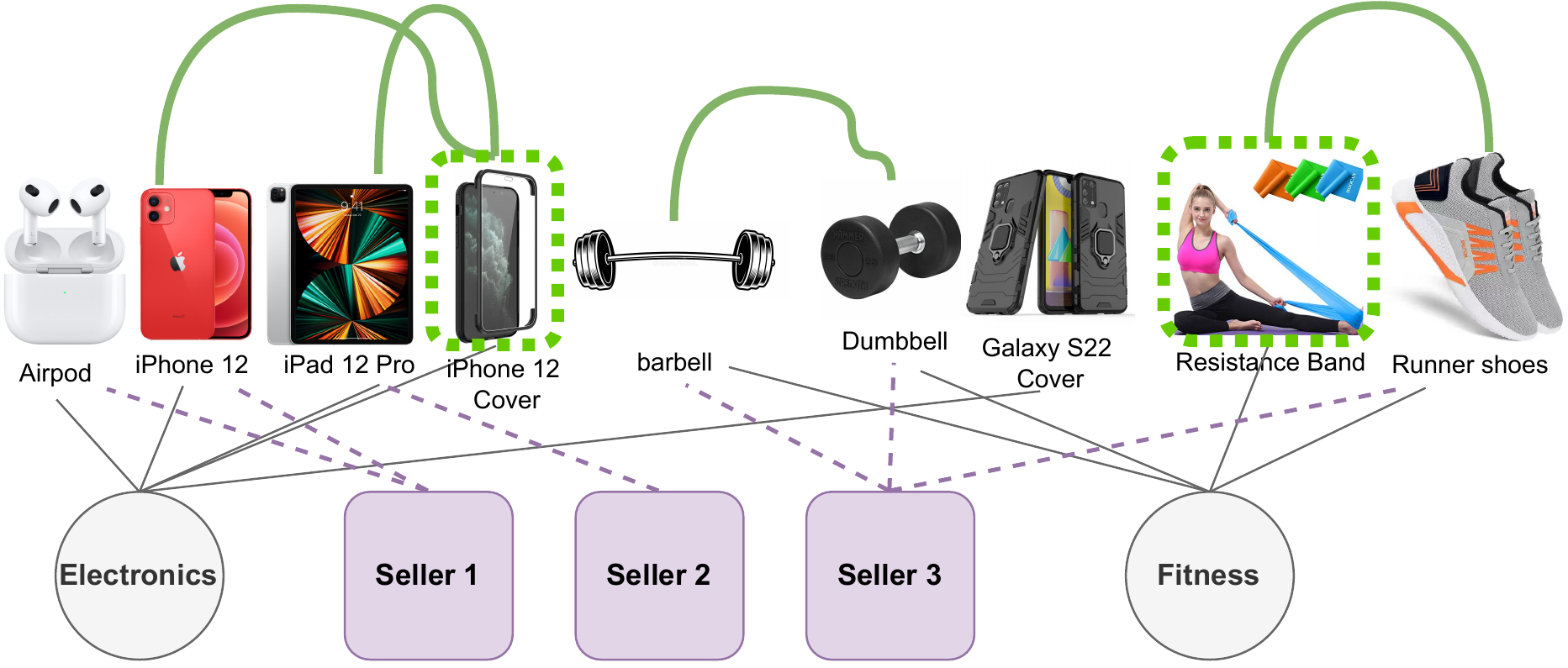}
         \caption{Edge threshold filter}
         \label{figure1c}
     \end{subfigure}
     \begin{subfigure}[b]{0.49\textwidth}
         \centering
         \includegraphics[width=0.99\textwidth]{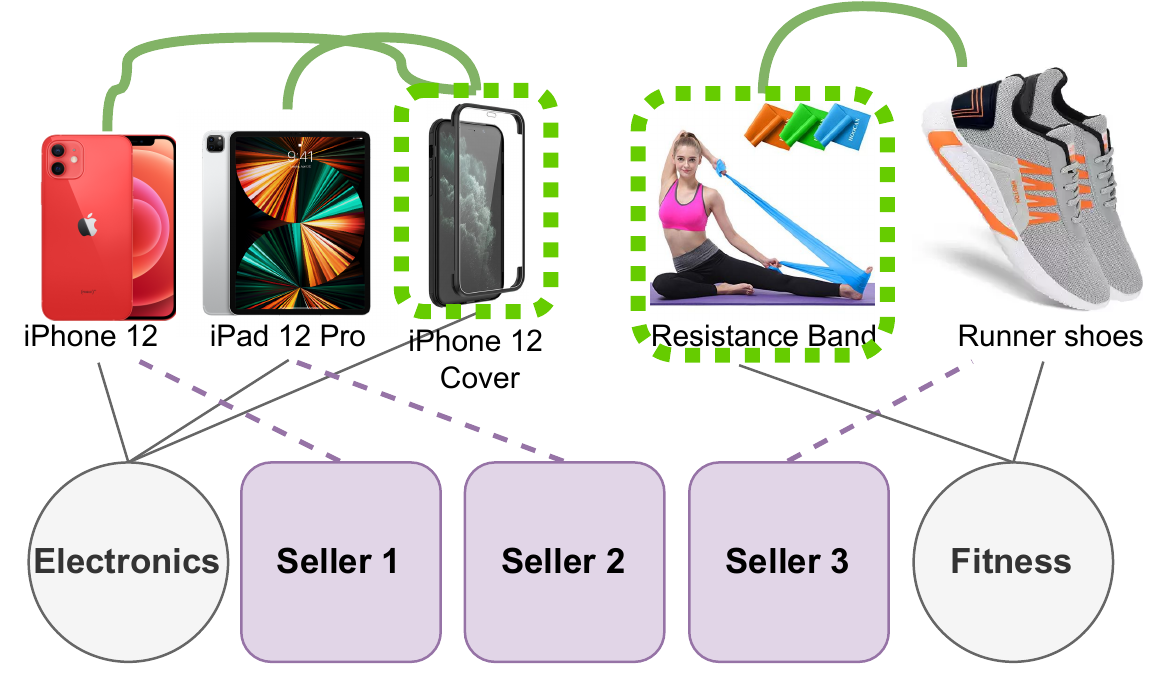}
         \caption{Remove products not connected to candidate set $\mathcal{P}_{\text{r}}$}
         \label{figure1d}
     \end{subfigure}
        \caption{Process to reduce edges and nodes in graph. Products surrounded in green dotted boxes \protect\includegraphics[height=\myMheight]{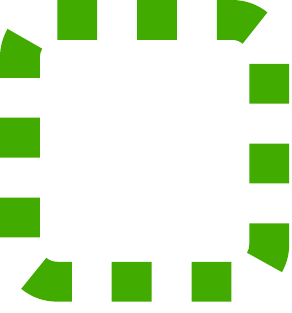} 
belong to our candidate set $\mathcal{P}_{\text{r}}$. (a) Initial Graph. (b) Remove customers and link products. (c) Remove product-product edges below a fixed threshold here $2$. (d) Remove products not connected to candidate set $\mathcal{P}_{\text{r}}$, forming our training product set $\mathcal{P}_{\text{t}} | \mathcal{P}_{\text{r}} \subset \mathcal{P}_{\text{t}}$.}
        \label{figure1}
        \Description{A woman and a girl in white dresses sit in an open car.}
\end{figure*}

We have a set of customers $\mathcal{C}\ (\geq 200M)$\footnote{M = million ($10^6$); K = thousand ($10^3$); B = Billion ($10^9$)} interacting with products $\mathcal{P}\ (\geq 200M)$ which belong to product categories $\mathcal{A}$, and sellers $\mathcal{S}\ (\geq 300K)$ selling these products\footnote{Numbers given here are for demonstrative purposes to give a sense of how our method down-samples the product graph. These are not actual business derived numbers of any e-commerce store.}. 
Customer-product interactions\footnote{We anonymise the customers for the product graph before using their interactions to build the graph, any customer data used in our system cannot be de-anonymised.} form edges $\mathbin{E^{\mathcal{C} \cup \mathcal{P}}}\ (\geq 5B)$ which are the largest edge set in our data, seller-product offerings form edges $\mathbin{E^{\mathcal{S} \cup \mathcal{P}}}\ (\sim 200M)$, and product to product category mappings form edges $\mathbin{E^{\mathcal{A} \cup \mathcal{P}}}\ (\sim 100M)$. 
Let $\mathcal{P_\text{r}} \subset \mathcal{P}$ be our candidate set of products from which we make recommendations which is $\sim 500K$. Since our recommendations are only for sellers, we removed all customer nodes $\mathcal{C}$ by linking any two products bought by same customer and only take those product to product links which occur over $500$ times in our dataset. 
We remove any product node not in 1-hop neighbourhood of any node in $\mathcal{P_\text{r}}$ (our candidate item set) and obtain a product set of candidate items $\mathcal{P_\text{r}}$ with their 1-hop neighbours from $\mathcal{P}$ forming our training product set $\mathcal{P_\text{t}} \mid \mathcal{P_\text{r}} \subset \mathcal{P_\text{t}} \subset \mathcal{P}$ and product to product edges $\mathbin{E^{\mathcal{P_\text{t}}}}$. 

Our final graph $\mathcal{G} = (\mathcal{V},\mathcal{E})$ is composed of nodes $\mathcal{V} = \mathcal{S} \cup \mathcal{P_\text{t}} \cup \mathcal{A}$ and edges $\mathcal{E} = \mathbin{E^{\mathcal{P_\text{t}}}} \cup \mathbin{E^{\mathcal{S} \cup \mathcal{P_\text{t}}}} \cup \mathbin{E^{\mathcal{A} \cup \mathcal{P_\text{t}}}}$. Our graph reduction mechanism enables us to reduce our graph size from $5B$ edges and $400M$ nodes to just $20M$ edges (250x lower) and $6M$ nodes (66x lower). For an example of this process see \sref{Figure}{figure1}, we provide the detailed steps of this approach in \sref{Appendix}{section:pg_extended}. \sref{Table}{table:graph_stat} provides details on approximate relative count of edges and nodes we obtain before and after filtering.

\enlargethispage{10pt}

\vspace*{-8pt}
\subsection{WAML architecture}
\label{subsection:waml}

\begin{figure*}[!t]
     \centering
     	 \begin{subfigure}[b]{0.49\textwidth}
	 	\centering
     	\begin{subfigure}[t]{0.5\textwidth}
         	\centering
         	\includegraphics[width=0.99\textwidth]{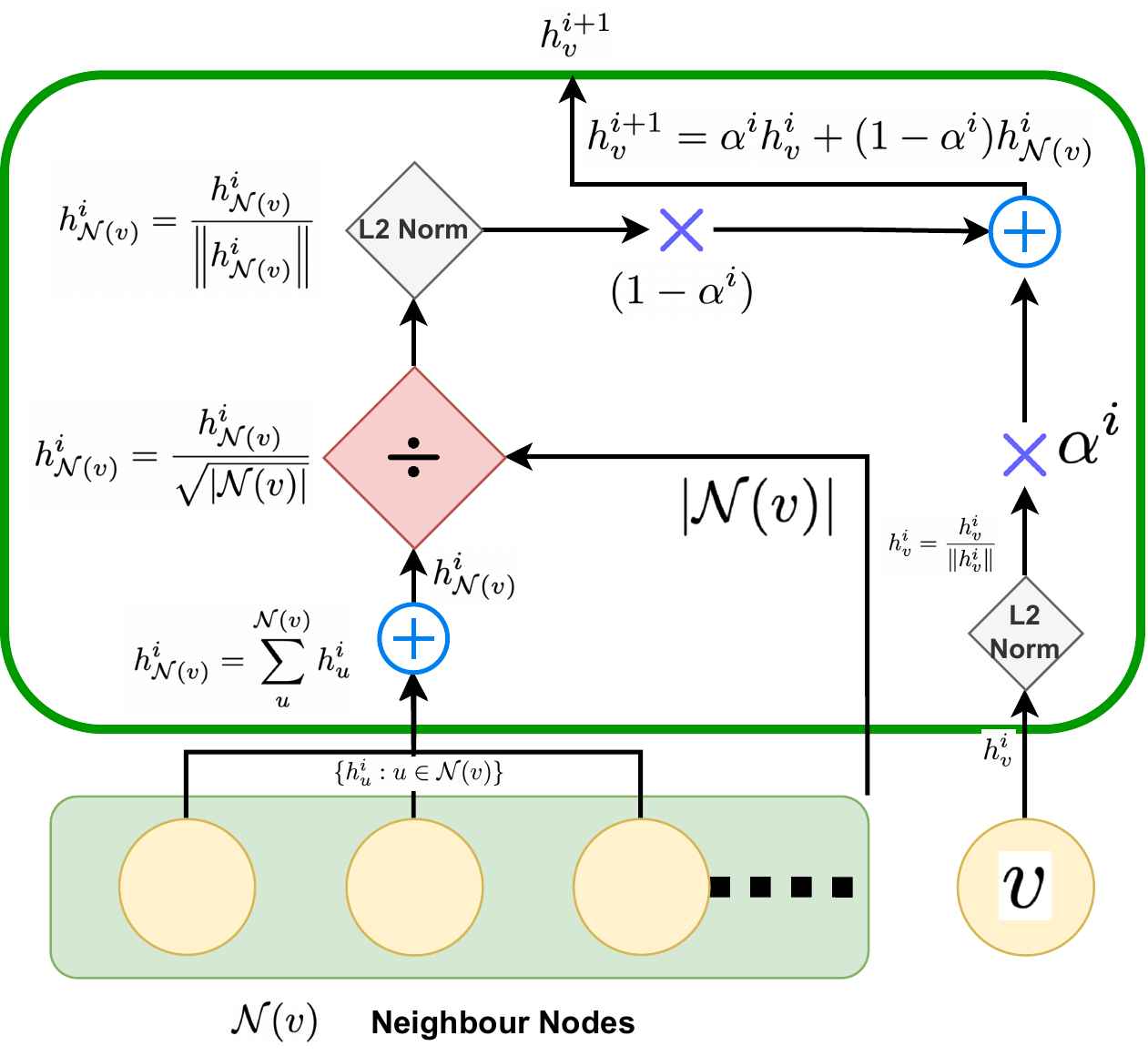}
         	\caption{WAML layer}
         	\label{figure2a}
     	\end{subfigure}
     	\begin{subfigure}[b]{0.5\textwidth}
         	\centering
         	\includegraphics[width=0.99\textwidth]{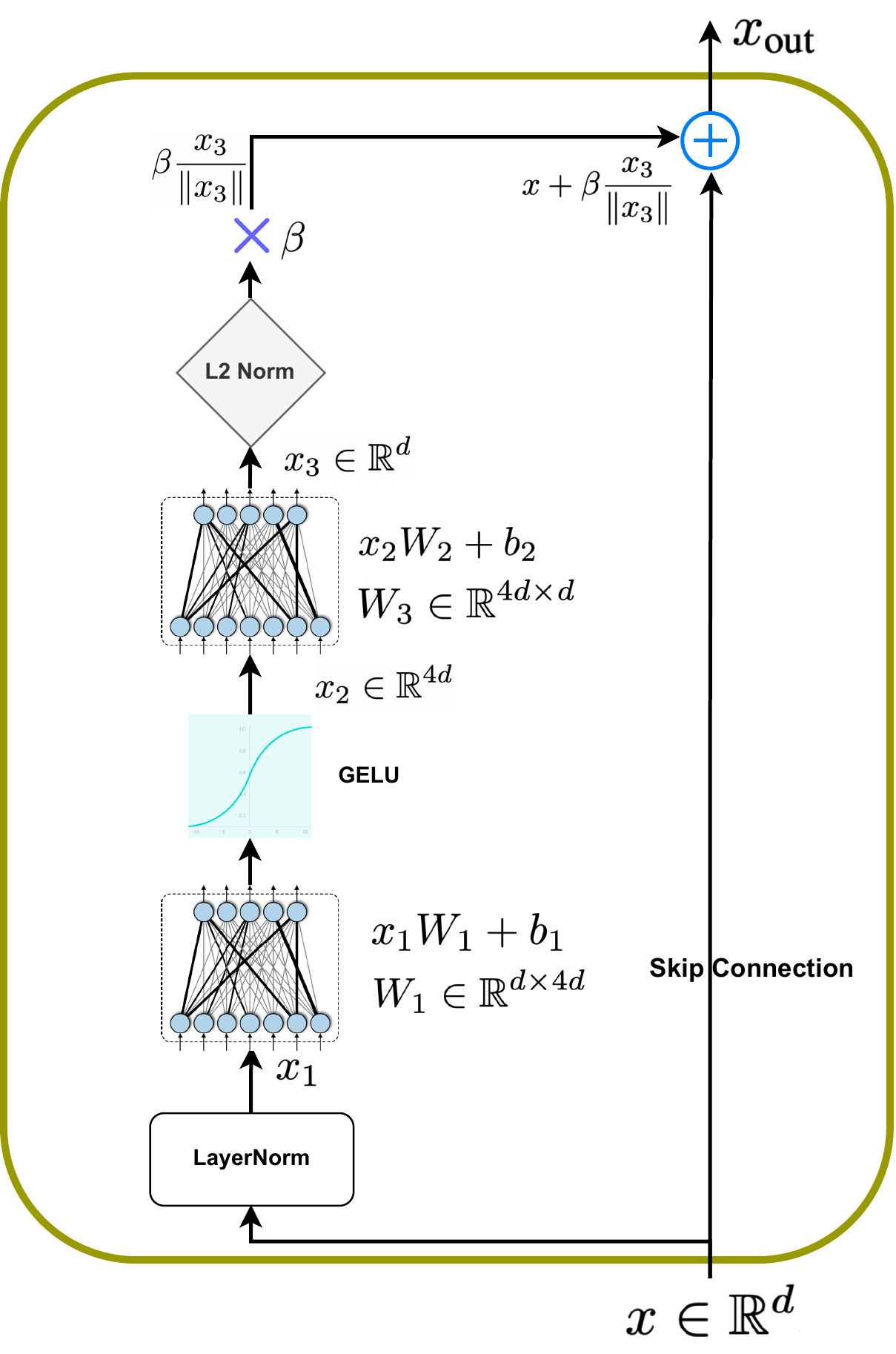}
         	\caption{Our custom BERT-FFN}
         	\label{figure2b}
     	\end{subfigure}
     \end{subfigure}
     \begin{subfigure}[b]{0.5\textwidth}
         \centering
         \includegraphics[width=0.95\textwidth]{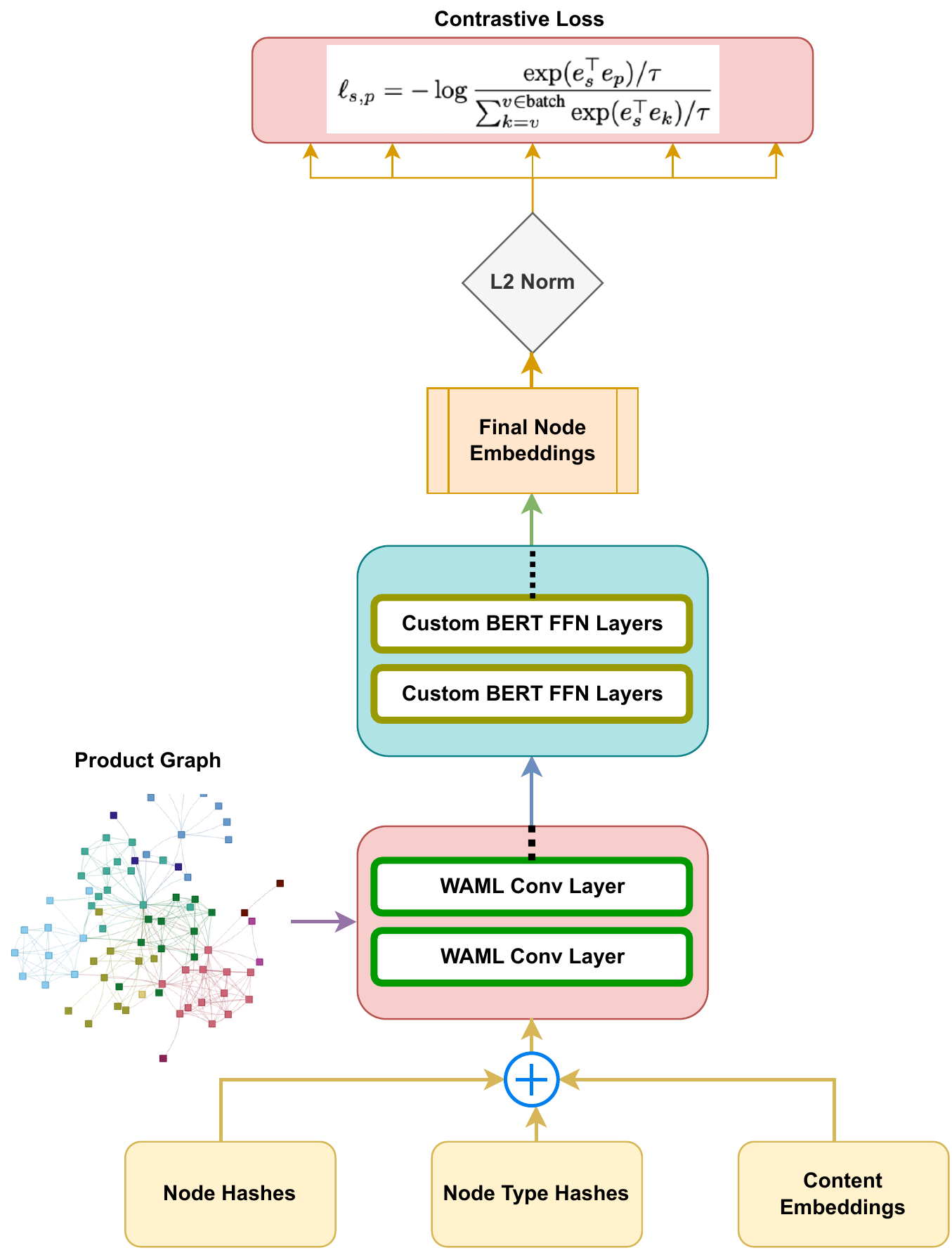}
         \caption{WAML Architecture}
         \label{figure2c}
     \end{subfigure}
     
        \caption{WAML architecture and it's components.}
        \label{figure2}
        \Description{A woman and a girl in white dresses sit in an open car.}
\end{figure*}

Each node $v \in \mathcal{V}$ in our graph is associated with real-valued attributes $x_v \in \mathbb{R}^{d}$ known as content features. For products $\mathcal{P}$ this is obtained by using BERT encoder on product title and description. For sellers $\mathcal{S}$ and product categories $\mathcal{A}$ nodes, we fill these with zeros. 
Our candidate products $\mathcal{P_\text{r}}$ have very less interactions with our sellers $\mathcal{S}$, see \sref{Table}{table:interactions}. As such our algorithm must be able to balance usage of content features with sparse collaborative signals from the graph. 
We denote neighbour nodes of a node $v \in \mathcal{V}$ as $\mathcal{N(\mathnormal{v})}$, each node's representation after a layer of graph convolution as $h^{i}_{v}$ where $i \in \{1, \dots, K\}$ and $K$ is the number of graph convolution layers in the model, node's neighbourhood representations as $\{h^{i}_{u} : u \in \mathcal{N(\mathnormal{v})}\}$. We list the differences between our method WAML and other graph based methods such as LightGCN and GAT in \sref{Table}{table:differences} in \sref{Appendix}{section:additional_results}.

\enlargethispage{11pt}

Our WAML architecture is depicted in \sref{Figure}{figure2}. Node representations on starting are a combination of three different embeddings, non-trainable node identifier hashes, node type identifier hashes, content embedding from BERT. Non-trainable node identifier hash $\text{HASH}(\text{Node Id}) \in \mathbb{R}^{d}$ replaces trainable per node embeddings, which scale well with increasing nodes in graph. We also incorporate node type hashes $\text{HASH}(\text{Node type}) \in \mathbb{R}^{d}$ to ensure WAML can treat each node type independently and process seller, product and product category nodes separately if needed. We use content embedding $x_v$ for all product nodes $\mathcal{P_\text{t}}$ from BERT output of each product's title and description while setting content embedding to zeros for sellers $\mathcal{S}$ and product categories $\mathcal{A}$. These embeddings before being fed into WAML layers are denoted by $h^{0}_{v} : v \in \mathcal{V}, h^{0}_{v} \in \mathbb{R}^{d}$ where $h^{0}_{v} = \text{HASH}(\text{Node Id}) + \text{HASH}(\text{Node type}) + x_v$.
Next, WAML layers, see \sref{Figure}{figure2a}, convolve over each node $v \in \mathcal{V}$ with the first layer input as $h^{0}_{v}$ and layer $i$ output as $h^{i}_{v}$. 
During each convolution a node $v$ combines it's own features $h^{i}_{v}$ with it's aggregated neighbour nodes features $h^{i}_{\mathcal{N(\mathnormal{v})}} = \sum_{u}^{\mathcal{N(\mathnormal{v})}} h^{i}_{u}$ using a weighted sum operation.
Both node features $h^{i}_{v}$ and aggregated neighbourhood features $h^{i}_{\mathcal{N(\mathnormal{v})}}$ are L$2$ normalized before they are combined. Each WAML layer $i$ has a parameter $\alpha^{i}$ which controls how much of neighbourhood features are integrated into the current node.
For $\forall v \in \mathcal{V}$ the WAML layer $i$ performs the following steps:\vspace*{-10pt}
\begin{align*}
h^{i}_{v} &= \frac{h^{i}_{v}}{\norm{h^{i}_{v}}} \\
h^{i}_{\mathcal{N(\mathnormal{v})}} &= \sum_{u}^{\mathcal{N(\mathnormal{v})}} h^{i}_{u} \\
h^{i}_{\mathcal{N(\mathnormal{v})}} &= \frac{h^{i}_{\mathcal{N(\mathnormal{v})}}}{\sqrt{\abs{\mathcal{N(\mathnormal{v})}}}} \\
h^{i}_{\mathcal{N(\mathnormal{v})}} &= \frac{h^{i}_{\mathcal{N(\mathnormal{v})}}}{\norm{h^{i}_{\mathcal{N(\mathnormal{v})}}}} \\
h^{i+1}_{v} &= \alpha^{i} h^{i}_{v} + (1 - \alpha^{i}) h^{i}_{\mathcal{N(\mathnormal{v})}}
\end{align*}

Our WAML stack is composed of $K$ parameter free layers with only a tunable $\alpha^{i}$ per layer $i$ and each WAML layer uses output of previous layer as input node features.
Tuning $\alpha^{i}$ allows us to ensure that $h^{i}_{v}$ is not diluted with neighbourhood features and content embeddings $x_v$ which were part of initial node features $h^{0}_{v}$ remain relevant after multiple WAML layers after graph structure has been incorporated into final node embedding $h^{K}_{v}$. In GCMC \citep{gcmc}, GAT \citep{gat} and GraphSAGE \citep{graphsage} the node and neighbourhood combination is learned with a neural network at each layer through back-propagation, this results in the graph structure taking high precedence over content embeddings, and overfitting to the graph structure present in training. LightGCN \citep{lightgcn} simply takes sum of neighbourhood nodes and uses that as next layer representation of node $v$ as $h^{i+1}_{v} = h^{i}_{\mathcal{N(\mathnormal{v})}}$. which dilutes node content embeddings with neighbourhood features without any control in our hands.

\begin{table*}[!t]
	\small
	\caption{Differences between WAML and other graph based methods.}
	\label{table:differences}
	
	\centering
	\begin{tabular}{lrrrr}
		\toprule
		
						     			& GAT 	& LightGCN & WAML \\
			\bt Architecture aspect 		&		   & 	  \\
		\midrule
			Node embeddings							& Trainable			 						& Trainable		   					& Hashing \citeauthor{hashingvectorizer}	  \\
			Node Type embeddings						& None				 						& None		   						& One-hot 	  \\
			Content embeddings						& Yes		 				 				& None		   						& Yes, from BERT 	  \\
			Self connection							& Yes				 						& None		   						& Weighted	  \\
			Neighbour aggregator						& Attention				 					& $\sum_{\mathcal{N(\mathnormal{v})}}{h^{i}_{u}}$		   	& $\alpha . h^{i}_{v} + (1 - \alpha) . h^{i}_{\mathcal{N(\mathnormal{v})}}$ 	  \\
			Trainable layer weights				& Yes				 						& None		   						& None 	  \\
			Layer-wise non-linearity					& RELU				 						& None		   						& None	  \\
			Training objective						& Rating prediction				 			& BPR \citeauthor{bpr}		   			& contrastive loss	  \\
			Regularization							& Dropout, L$2$					 			& L$2$		   							& L$2$, Dropout 	  \\
			Negative sampling						& None				 						& None	   							& same mini-batch	  \\
			Normalization per layer					& $\frac{h^{i}_{v}}{\sqrt{\abs{\mathcal{N(\mathnormal{v})}}}}$	&	$\frac{h^{i}_{v}}{\sqrt{\abs{\mathcal{N(\mathnormal{v})}}}}$   							& $\frac{h^{i}_{v}}{\sqrt{\abs{\mathcal{N(\mathnormal{v})}}}}$, L2	  \\
			DNN after Graph-conv						& bilinear decoder				 			& None		   							& BERT-FFN	  \\
			Output normalization						& None			 							& None		   							& L$2$	  \\
			
		\bottomrule
	\end{tabular}
\end{table*}

\subsection{WAML training with Contrastive loss}
\label{subsection:cncf}
After we obtain node embeddings $h^{K}_{v}: \forall v \in \mathcal{V}$ from WAML stack as described in \sref{Section}{subsection:waml}, we propagate them through a customised BERT-FFN network inspired by \citet[Section~3.3]{attention}, to obtain final node embeddings and then calculate the loss on these node embeddings. Our method uses a contrastive loss while previous methods \citep{ncf} simply predicted implicit ratings.
We pass $h^{K}_{v}$ through customised BERT-FFN layer stack, see \sref{Figure}{figure2b}.
Each BERT-FFN layer $j \in J$, where $J = 3$ is number of BERT-FFN layers, takes input from previous layer as $e^{j}_{v}: \forall v \in \mathcal{V}$ and produces $e^{j+1}_{v}: \forall v \in \mathcal{V}$ for input to next layer. Final layer outputs of node representations $e^{J}_{v}: \forall v \in \mathcal{V}$ are L2-normed and then passed onto the loss function. L2-normalizing the output vectors as $e_{v} = \frac{e^{J}_{v}}{\norm{e^{J}_{v}}}$, ensures the loss is minimized in the cosine similarity space, these vectors can be efficiently queried using any off the shelf K-nearest neighbour search engine. Each BERT-FFN layer does the following for an input vector $x \in \mathbb{R}^d$:
\begin{align*}
	x_{1} &= \mathrm{LayerNorm}(x)\\
	x_{2} &= \mathrm{GELU}(x_{1}W_{1} + b_{1}) \mid W_{1} \in \mathbb{R}^{d\times4d}, x_{2} \in \mathbb{R}^{4d}\\
	x_{3} &= x_{2}W_{2} + b_{2} \mid W_{3} \in \mathbb{R}^{4d\times d}, x_{3} \in \mathbb{R}^{d}\\
	x_{\text{out}} &= x + \beta \frac{x_{3}}{\norm{x_{3}}}
\end{align*}

In last step BERT-FFN adds the input $x$ to it's internal representation $x_{3}$ after normalization and multiplication by $\beta$ as $x_{\text{out}} = x + \beta \frac{x_{3}}{\norm{x_{3}}}$. Normalization followed by $\beta$ scaling ensures that the input $x$ is only slightly changed, which is essential to preserve content attributes $x_v$ which were present in WAML stack's input $h^{0}_{v}$ and were propagated to BERT-FFN stack as input $e^{0}_{v} = h^{K}_{v}$. 

Inspired by \citet[Section~2.1]{simclr}, our contrastive loss function aims to minimize the distance between node pairs  $(s, p) \in \mathbin{E^{\mathcal{S} \cup \mathcal{P_\text{t}}}}$ where $p \in \mathcal{P_\text{t}}$ and $s \in \mathcal{S}$, while maximizing distance between node pairs $(s, p) \notin \mathbin{E^{\mathcal{S} \cup \mathcal{P_\text{t}}}}$, intuitively we minimize distance between node pairs belonging to seller-product edges $\mathbin{E^{\mathcal{S} \cup \mathcal{P_\text{t}}}}$ to bring seller nodes close to their linked product nodes, while we maximize distance between seller and product nodes which are not linked in the graph. 
We randomly sample a minibatch of $N$ seller-product edges $(s, p) \in \mathbin{E^{\mathcal{S} \cup \mathcal{P_\text{t}}}}$ and perform contrastive matching on seller-product edges, while using the remaining nodes $2(N-1)$ in batch as negative examples. The loss function for a positive seller-product pair $(s, p)$ whose final representations are $e_{s}$ and $e_{p}$ is defined as

\begin{equation}
	\label{eq:loss}
	\ell_{s,p}=-\log\frac{\exp(e_{s}^{\top}e_{p})/\tau}{\sum_{k=v}^{v \in \text{batch}}  \exp(e_{s}^{\top}e_{k})/\tau }
\end{equation}
where $\tau=0.1$ is a temperature parameter. We take mean of this loss function over all edges $(s, p)$ in our mini-batch as our overall loss for the batch and train WAML using back propagation. We list various architectural differences between WAML and other graph based methods in \sref{Table}{table:differences}.

\section{Results and Ablations}
\label{section:results}
We test our WAML algorithm on our productionized usecase of seller-product recommendations. The process of dataset creation is mentioned in \sref{Section}{section:pg} and the dataset statistics are covered in \sref{Table}{table:graph_stat}. Baseline and ablation results are based on the product graph created in \sref{Section}{section:pg}, while we provide full product graph based results only on our final architecture. We use Recall@100 metric which determines the absolute retrieval capability for the first 100 results. For performance comparison we consider LightGCN + Content features as our primary baseline. While starting with our experiments we trained a \textbf{base} model and then made changes from base model to build our WAML model, we see the below differences to characterize our \textbf{base} architecture vs our final WAML architecture:
\begin{enumerate}
	\item We start with trainable node embeddings and don't use node id hashes or node type hashes. We also exclude content features $x_v$.
	\item We perform simple addition of node and neighbour embeddings with $\alpha^{i}=0.5$.
	\item No L2 normalization anywhere, normalize neighbour embeddings by $h^{i}_{\mathcal{N(\mathnormal{v})}} = \frac{h^{i}_{\mathcal{N(\mathnormal{v})}}}{\sqrt{\abs{\mathcal{N(\mathnormal{v})}}}}$.
	\item No BERT-FFN layer after WAML, pass WAML output to loss function.
	\item Triplet loss as our initial loss function.
\end{enumerate}

\begin{table*}[!t]
	\small
	\centering

	\begin{tabular}{lccccc}
		\toprule
				  & \bt Compute & \bt Max Memory & \bt Params  & \bt Recall@100   & {\bt Recall@100}  \\
		\bt Model & (vs Ours)   & (vs Ours)  &           (All / Network)   &     & $(\mathrm{Gain}\%)$  \\
		\midrule
		& & &{\bt Baselines} & &\\
		SVD 	& 2.2x 				& 0.4x  & 1.9B / 0.0M   & 0.0151  & ${92.10\%  \downarrow}$ \\
		NCF + Content & 7.5x    			& 5.2x  & 1.9B / 3.2M   & 0.0626  & ${67.26\%  \downarrow}$ \\
		GCMC       & 15x 			& 9.0x  & 1.9B / 4.3M   	& 0.1229  & ${35.72\%  \downarrow}$ \\
		GAT + DropoutNet       	& 26x 	& 12.1x & 1.9B / 8.2M  & 0.1630  & ${14.74\%  \downarrow}$ \\
		LightGCN + Content       		& 7.1x 	& 8.2x  & 1.9B / 1.1M		& 0.1912  & ${00.00\%  \uparrow}$ \\
		
		\midrule
		& & &{\bt Ours} & &\\
		
		Base        & 5.8x 			& 6.5x  & 1.9B / 0.0M     	& 0.1391  & ${ 27.25\%  \downarrow}$\\
		+ Content       & 7.1x 		& 7.8x  & 1.9B / 1.1M  & 0.1871  & ${ 02.14\%  \downarrow}$ \\
		- Node embeddings   & 1.1x 	& 0.7x  & 1.1M / 1.1M  & 0.1426  & ${ 25.42\%  \downarrow}$ \\
		+ Node Id and type hash     & 1.3x & 0.7x  & 1.1M / 1.1M    & 0.1607  & ${ 15.95\%  \downarrow}$ \\
		+ L2-Norm WAML & 1.1x 		& 0.7x  & 1.1M / 1.1M     	& 0.1721  & ${09.98\%  \downarrow}$ \\
		+ tune $\alpha^{i}$ & 1.1x 	& 0.7x  & 1.1M / 1.1M  & 0.2211  & ${ 15.63\%  \uparrow}$ \\
		+ Simple FFN & 1.4x  		& 0.8x  & 3.2M / 3.2M	    & 0.2574  & ${ 34.62\%  \uparrow}$ \\
		+ BERT-FFN & 1.8x  			& 1.0x  & 6.6M / 6.6M	    & 0.2720  & ${ 42.25\%  \uparrow}$ \\
		+ L2-Norm BERT-FFN & 1.6x  	& 1.0x  & 6.6M / 6.6M 	& 0.2751  & ${ 43.88\%  \uparrow}$ \\
		+ tune $\beta$ &  1.6x 		& 1.0x  & 6.6M / 6.6M  & 0.2835  & ${ 48.27\%  \uparrow}$ \\
		+ Contrastive loss &   1.2x & 1.0x  & 6.6M / 6.6M     	& 0.3136  & ${ 64.01\%  \uparrow}$ \\
		+ L2-Norm output &    1.0x  & 1.0x  & 6.6M / 6.6M 	& 0.3179  & \bm{${ 66.27\%  \uparrow}$} \\
		+ full dataset  & 45x 		&1.0x  	& 6.6M / 6.6M 		& 0.4051  & \bm{${ 111.8\%  \uparrow}$} \\
		\bottomrule
	\end{tabular}
	\caption{Results of various algorithms on our use-case. See \sref{Table}{table:main_results_extended} for more detailed results.}
	\label{table:main_results}
\vspace*{-11pt}
\end{table*}

In \sref{Table}{table:main_results}, we present our results with ablations. Compute comparison is based on wall time on p3.2x large AWS machine ($1$ Nvidia V100 16GB). Memory comparison is based on maximum memory used during its execution, divided by batch size. Trainable params for the entire neural network and for non-embedding GNN/FFN parts are reported side by side, for SVD only params are the parameters of the factorized user and item matrices. The SVD implementation is taken from scikit-surprise python library \citep{surprise} which has efficient C++ implementation of SVD. The remaining algorithms were implemented in python and pytorch \citep{pytorch}. Content features are integrated by processing them first with a simple feed forward network. We set node embeddings to $256$ dimensions and content features to $256$ dimensions and train using AdamW \citep{adamw} optimiser with a learning rate of $0.0001$, using a train-validation-test split to minimize loss on validation split and then report results on test split.

Few observations that can be inferred from our results:
\begin{enumerate}
	\item The majority of parameters is composed of user and product embedding matrices. We have $1.86B$ parameters for embeddings only. When we drop node embeddings and use node id hash \citep{hashingvectorizer} we see a drastic drop in both compute and memory usage, and a performance degradation in Recall@100 from $0.1820$ to $0.1486$. Node embeddings encode graph structure, addition of node id and type hash helps us to encode graph structure with no params and less compute. Unlike previous methods like GraphSAGE \citep{graphsage} and PinSAGE \citep{pinsage} which are purely inductive and skip any node identifiers for scaling, we notice that non-trainable node identifier hashes increase WAML's performance while sacrificing pure inductive capabilities which are not important to our usecase.
	\item Adding L2-norm in our architecture reduces our overall compute due to faster convergence. L2-Norms are inexpensive operation on GPU which stabilise the network training and improve performance minorly.
	\item Node type hashes increase performance minorly, in our case since we only have three node types but we believe with usecases having higher variety of node types we will observe more gains from node type hashes.
	\item Tuning $\alpha^{i}$ for each layer leads to huge jump in performance, a gain of $40\%$ over \textbf{base} and $15\%$ over LightGCN. Our network has $5$ WAML layers and their $\alpha^{i} = [0.4, 0.45, 0.5, 0.6, 0.7],$ $  i \in \{1, \dots, 5\}$. $\alpha^{i} \leq 0.5 \mid i \leq 3$ implying that neighbourhood information is necessary in early layers, but $\alpha^{i} > 0.5 \mid i > 3$ implying node's own information is important towards the end WAML layers. Being able to tune the proportion of neighbourhood vs node's own information per layer enables us to tilt the algorithm towards content features and solve our soft cold start issue.
	\item Contrastive loss decreases our training time compared to triplet loss, since in triplet loss one node is only compared to one positive and one negative sample, but in case of contrastive loss we can use the entire batch of nodes as negative examples. This leads to faster convergence while boosting recall by $11\%$ ($0.2835$ to $0.3136$).
	\item Graph networks like GAT exploit graph structure and their Recall@100 is good in collaborative domain, but in our case they de-emphasise content features and provide no control over how content features can be used, resulting in low performance and high compute. 
	\item LightGCN with content features outperforms GAT since content features aren't diluted through deep neural network in LightGCN, as it uses a simple addition to aggregate neighbourhood features. NCF and NCF + content both don't exploit graph structure and as a result perform lowest while still using significant compute.
	\item Using the non-filtered full dataset provides modest recall improvement of $27\%$ ($0.3179$ to $0.4051$) over the filtered dataset  with our WAML method but also results in massively increased compute by \bm{${\color{black} 45\text{x}}$} for training which proves the efficacy of our filtering method in \sref{Section}{section:pg}. 
\end{enumerate}

\section{Conclusion}
\label{section:conclusion}
We encountered a novel recommendation system problem where we have a perpetual soft item cold start issue for all candidate recommendation items. Our items/products always have very few interactions and once they gather enough interactions they are removed from the candidate set. We recognised that e-commerce recommendation system problems can be viewed as a link prediction problem in a partially observed dynamic graph.
Traditional algorithms like collaborative filtering with matrix factorization and more modern algorithms such as graph neural networks are meant to work where items have high number of interactions with content features used for only assisting the CF algorithm. Content based approaches on the other hand completely rely on content features and ignore the product graph. Simultaneously, it has been observed that graph neural networks are low pass filters on the graph data structure \citep{lowpass}, and for recommendation systems, feature transformations and non-linear activation on each graph neural network layer has no positive effect on collaborative filtering \citep{lightgcn}. These findings correlate with our findings and show why complex networks like GAT and GCMC underperform compared to WAML and LightGCN.

A controlled approach to combining content with product graph performs best in our domain. We also notice the recall to compute trade-off in modern graph algorithms such as GAT and GCMC is sub-par for recommendation systems, these algorithms are more suited for node and graph classification/regression tasks \citep{lightgcn}. LightGCN, a graph algorithm especially enhanced for recommendation systems provides a good compute balance but still fails to scale to production use cases and solve our soft cold start issue. We propose several changes, removal of node embeddings, graph reduction mechanism and controlled combination between node and neighbourhood embeddings through our WAML method in \sref{Sec.}{section:method} which enable us to scale and provide high recall for our use-case. Our innovative graph reduction mechanism reduces product graph size from $\geq 5B$ edges, $400M$ nodes to just $20M$ (250x lower) edges, $6M$ (66x lower) nodes\footnote{Numbers given here are for demonstrative purposes to give a sense of how our method down-samples the product graph. These are not actual business derived numbers of any e-commerce store.}. , resulting in 45x reduction in compute while retaining $79.5\%$ of maximum recall, see \sref{Table}{table:main_results}. On the same graph size our WAML method provides a gain of \bm{${\color{black} 66.27\%}$} over next best method by LightGCN.


\bibliographystyle{ACM-Reference-Format}
\bibliography{sample-base}

\appendix
\section{Building the Product Graph}
\label{section:pg_extended}

\begin{figure*}[!t]
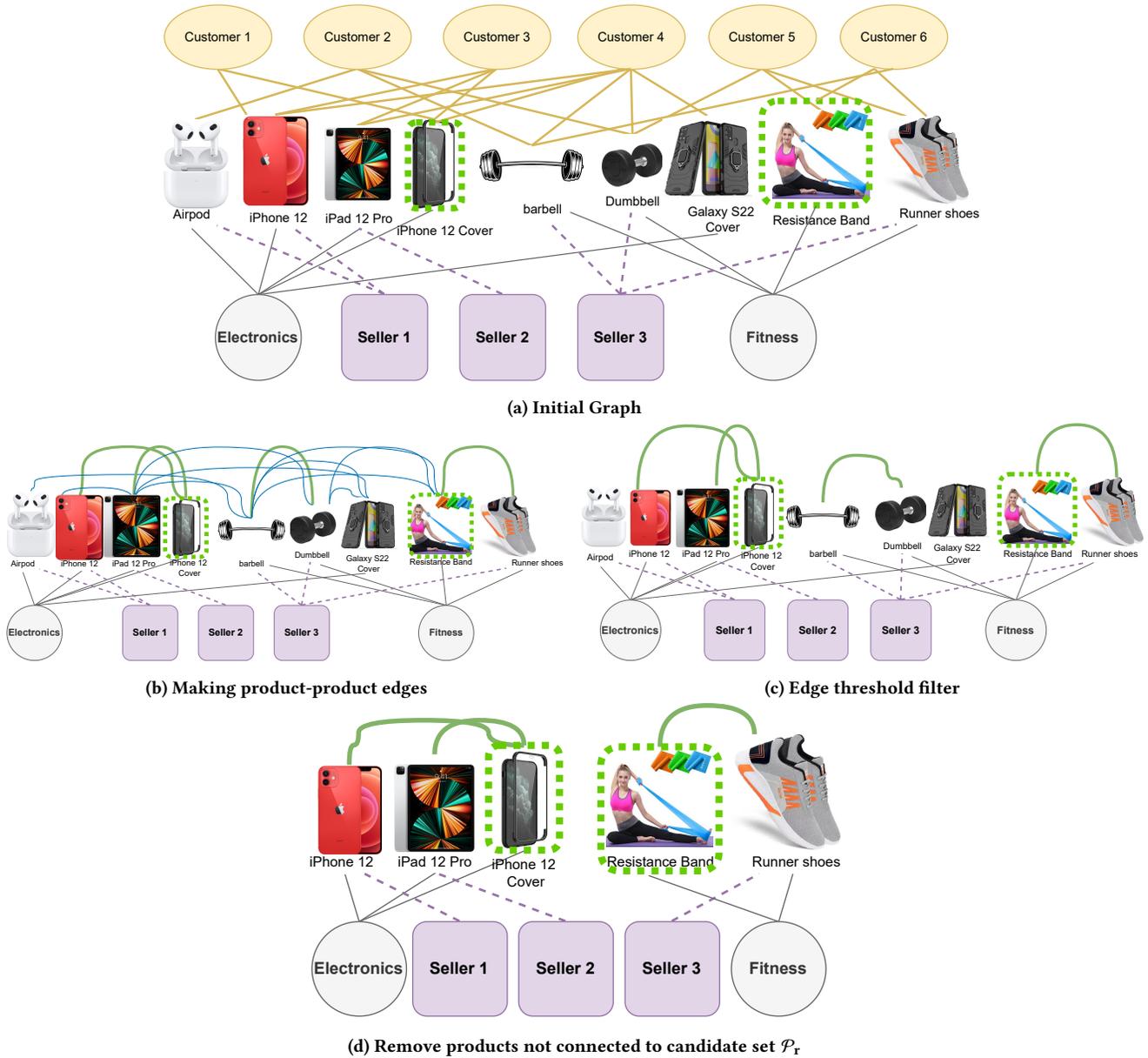

     \centering
     \begin{subfigure}[b]{0.75\textwidth}
         \centering
         \includegraphics[width=0.95\textwidth]{product_filtering-initial-drawio-v2.pdf}
         \caption{Initial Graph}
         \label{figure3a}
     \end{subfigure}
     \hfill
     \begin{subfigure}[b]{0.49\textwidth}
         \centering
         \includegraphics[width=0.99\textwidth]{product_filtering-step-1-drawio-v2.pdf}
         \caption{Making product-product edges}
         \label{figure3b}
     \end{subfigure}
     \begin{subfigure}[b]{0.49\textwidth}
         \centering
         \includegraphics[width=0.99\textwidth]{product_filtering-step-2-drawio-v2.pdf}
         \caption{Edge threshold filter}
         \label{figure3c}
     \end{subfigure}
     \begin{subfigure}[b]{0.5\textwidth}
         \centering
         \includegraphics[width=0.95\textwidth]{product_filtering-step-3-drawio-v2.pdf}
         \caption{Remove products not connected to candidate set $\mathcal{P}_{\text{r}}$}
         \label{figure3d}
     \end{subfigure}
        \caption{Process to reduce edges and nodes in graph. Products surrounded in green dotted boxes \protect\includegraphics[height=\myMheight]{product_filtering-green-box.pdf} 
belong to our candidate set $\mathcal{P}_{\text{r}}$. (a) Initial Graph with all customers $\mathcal{C}$ and products $\mathcal{P}$. (b) Remove customers and connect products interacted by same customer. (c) Remove product-product edges below a fixed threshold, here threshold $= 2$. (d) Remove products not connected to candidate set $\mathcal{P}_{\text{r}}$, remaining products form our training product set $\mathcal{P}_{\text{t}} | \mathcal{P}_{\text{r}} \subset \mathcal{P}_{\text{t}}$.
        }
        \label{figure3}
\end{figure*}

\begin{table*}[!t]
	\small
	\caption{Graph edge and node counts before and after our filtering technique.}
	\label{table:graph_stat}
	
	\centering
	\begin{tabular}{llr}
		\toprule
						     
			Symbol &	Description	     &  Approximate Count\\
		\midrule
			$\mathcal{S}$ & Sellers                                                                             & $1x$ \\
			$\mathcal{C}$ & Customers                                                                           & $1\times10^3x$ \\
			$\mathcal{P}$ & Products                                                                            & $1\times10^3x$ \\
			$\mathcal{A}$ & Product categories                                                                  & $\leq 100$ \\
			$\mathcal{P}_{\text{r}}$ & Candidate Products                                                         & $5x$ \\
			$\mathcal{P}_{\text{t}}$ & Training product set                                                       & $30x$ \\
			
			\midrule
			$\mathbin{E^{\mathcal{C} \cup \mathcal{P}}}$ & Customer-Product Edges                               & $2\times10^4x$ \\
			$\mathbin{E^{\mathcal{P}}}$ & Unfiltered product-product edges                                      & $1 \times 10^4x$ \\
			$\mathbin{E^{\mathcal{S} \cup \mathcal{P}}}$ & Unfiltered seller-product edges                      & $2\times10^2x$ \\
			$\mathbin{E^{\mathcal{A} \cup \mathcal{P}}}$ & Unfiltered product to product categories edges       & $1\times10^2x$ \\
			$\mathbin{E^{\mathcal{P}_{\text{t}}}}$ & Filtered product-product edges                               & $80x$ \\
			$\mathbin{E^{\mathcal{S} \cup \mathcal{P}_{\text{r}}}}$ & Seller to candidate product edges           & $2.2x$ \\
			$\mathbin{E^{\mathcal{S} \cup \mathcal{P}_{\text{t}}}}$ & Seller to filtered product edges            & $10x$ \\
			$\mathbin{E^{\mathcal{A} \cup \mathcal{P}_{\text{t}}}}$ & Filtered product to product categories edges & $30x$ \\
			\midrule
			$\mathcal{V} = \mathcal{S} \cup \mathcal{P}_{\text{t}} \cup \mathcal{A}$ & Final Graph Nodes          & $31x$ \\
			$\mathcal{E} = \mathbin{E^{\mathcal{P}_{\text{t}}}} \cup \mathbin{E^{\mathcal{S} \cup \mathcal{P}_{\text{t}}}} \cup \mathbin{E^{\mathcal{A} \cup \mathcal{P}_{\text{t}}}}$ & Final Graph Edges & $120x$ \\

		\bottomrule
	\end{tabular}
\end{table*}

Within any e-commerce store we have a set of customers $\mathcal{C}$, products $\mathcal{P}$ and sellers $\mathcal{S}$ selling these products. 
Let $\abs{\mathcal{C}}$ be the total number of customers, $\abs{\mathcal{P}}$ be total products and $\abs{\mathcal{S}}$ be total sellers. 
A customer $c_i \in \mathcal{C}$ with $i \in \{1, \dots, \abs{\mathcal{C}}\}$ interacts with a product $p_j \in \mathcal{P}$ with $j \in \{1, \dots, \abs{\mathcal{P}}\}$ through clicks, add to cart, views and purchases forming an edge tuple as $(c_i, p_j) \in \mathbin{E^{\mathcal{C} \cup \mathcal{P}}}$ where $\mathbin{E^{\mathcal{C} \cup \mathcal{P}}}$ denotes all customer-product edges. 
Similarly a seller $s_k \in \mathcal{S}$ with $k \in \{1, \dots, \abs{\mathcal{S}}\}$ interacts with product $p_j$ by listing/offering the product for sale in the store forming an edge tuple as $(s_k, p_j) \in \mathbin{E^{\mathcal{S} \cup \mathcal{P}}}$ where $\mathbin{E^{\mathcal{S} \cup \mathcal{P}}}$ denotes all seller-product edges. 

Apart from customer-product and seller-product edges, each product is also associated with its corresponding product category $a_m \in \mathcal{A}$ with $m \in \{1, \dots, \abs{\mathcal{A}}\}$, where $\mathcal{A}$ denotes all product categories in the store with $\abs{\mathcal{A}}$ total categories. This product $p_j$ to product-category $a_m$ mapping produces edge tuples as $(p_j, a_m) \in \mathbin{E^{\mathcal{A} \cup \mathcal{P}}}$ where $\mathbin{E^{\mathcal{A} \cup \mathcal{P}}}$ denotes all product to product category edges. As a result, the overall set of entities and their interactions can be represented as an undirected graph as shown in \sref{Figure}{figure3a}.

After obtaining customers $\mathcal{C}$, products $\mathcal{P}$, sellers $\mathcal{S}$ and customer-product edges $\mathbin{E^{\mathcal{C} \cup \mathcal{P}}}$, seller-product edges $\mathbin{E^{\mathcal{S} \cup \mathcal{P}}}$ and product to product category edges $\mathbin{E^{\mathcal{A} \cup \mathcal{P}}}$. While the count of set of products $\mathcal{P}$ is $\abs{\mathcal{P}} > 200$ Million, the count $\abs{\mathcal{P_\text{r}}}$ of our candidate set of products $\mathcal{P_\text{r}} \subset \mathcal{P}$ from which we recommend to sellers is $\abs{\mathcal{P_\text{r}}} \sim 500K$. Also our recommendation system is intended for sellers only. 
With these two observations, we decided that we can remove all customers $\mathcal{C}$ nodes as well as any product node not in 1-hop neighbourhood of any node in $\mathcal{P_\text{r}}$.
Specifically we followed the below steps for node and edge filtering, an example of such process can be seen in \sref{Figure}{figure3}.
\begin{enumerate}
    \item For any two customer-product edge $(c_i, p_j)$ and $(c_i, p_k)$ where customer $c_i$ is common, we add a new product-product edge $(p_j, p_k) \in \mathbin{E^{\mathcal{P}}}$ where $\mathbin{E^{\mathcal{P}}}$ denote all product-product edges. Then We remove all customer nodes $\mathcal{C}$ and all customer-product edges $\mathbin{E^{\mathcal{C} \cup \mathcal{P}}}$
    \item There can be multiple edges between two products $p_i$ and $p_k$ if both products were interacted with together by multiple customers. For example, $p_i$ can be ``iPhone 12 Pro 256 GB'' and $p_j$ can be ``iPhone 12 Pro case cover''. Some irrelevant product-product pairs may also be included with low frequency, to capture only relevant pairs we apply a threshold of $200$ and drop any pair $(p_j, p_k)$ which has occurred less times than our threshold.
    \item We consider products in our candidate set $\mathcal{P_\text{r}} \subset \mathcal{P}$ and keep only those product-product edges $(p_j, p_k)$ where $p_j \in \mathcal{P_\text{r}}$ or $p_k \in \mathcal{P_\text{r}}$. Intuitively, one of the product node in a product-product edge must be in our smaller candidate set of recommendations. This gives us a new product subset which includes the candidate set $\mathcal{P_\text{r}}$ and few other products from $\mathcal{P}$, which we refer as our training product set $\mathcal{P_\text{t}} \mid \mathcal{P_\text{r}} \subset \mathcal{P_\text{t}} \subset \mathcal{P}$ with its count $\abs{\mathcal{P_\text{t}}} : \abs{\mathcal{P_\text{r}}} \ll \abs{\mathcal{P_\text{t}}} \ll \abs{\mathcal{P}}$. This reduces the count of product-product edges giving us $\mathbin{E^{\mathcal{P_\text{t}}}} \subset \mathbin{E^{\mathcal{P}}}$ as our final product-product edges.
    \item We keep all sellers $\mathcal{S}$ and any seller-product edges $(s_k, p_j)$ where $p_j \in \mathcal{P_\text{t}}$ with $j \in \{1, \dots, \abs{\mathcal{P_\text{t}}}\}$ to form new set of seller-product edges $\mathbin{E^{\mathcal{S} \cup \mathcal{P_\text{t}}}} \subset \mathbin{E^{\mathcal{S} \cup \mathcal{P}}}$. Note that the number of seller to candidate product edges $\mathbin{E^{\mathcal{S} \cup \mathcal{P_\text{t}}}}$  are far less since $\abs{\mathcal{P_\text{r}}} \ll \abs{\mathcal{P_\text{t}}}$. Refer \sref{Table}{table:interactions} to see how few interactions products from our candidate set $\mathcal{P_\text{r}}$ get.
    \item we keep all product to product category edges $(p_j, a_m)$ where $p_j \in \mathcal{P_\text{t}}$ with $j \in \{1, \dots, \abs{\mathcal{P_\text{t}}}\}$ to form new set of product to product category edges  $\mathbin{E^{\mathcal{A} \cup \mathcal{P_\text{t}}}} \subset \mathbin{E^{\mathcal{A} \cup \mathcal{P}}}$.
\end{enumerate}

Our final graph $\mathcal{G} = (\mathcal{V},\mathcal{E})$ is composed of nodes $\mathcal{V} = \mathcal{S} \cup \mathcal{P_\text{t}} \cup \mathcal{A}$ and edges $\mathcal{E} = \mathbin{E^{\mathcal{P_\text{t}}}} \cup \mathbin{E^{\mathcal{S} \cup \mathcal{P_\text{t}}}} \cup \mathbin{E^{\mathcal{A} \cup \mathcal{P_\text{t}}}}$, similar to \sref{Figure}{figure3d}. When we describe our WAML algorithm, we refer to this final graph $\mathcal{G}(\mathcal{V},\mathcal{E})$ and train all our experiments on this graph. Note that our product graph has more products $\mathcal{P_\text{t}}$ than the candidate set $\mathcal{P_\text{r}}$ which helps in creating links between products from $\mathcal{P_\text{r}}$ and sellers $\mathcal{S}$ since products from candidate set themselves have very less direct links with sellers. \sref{Table}{table:graph_stat} provides details on approximate relative count of edges and nodes we obtain before and after filtering.

\section{Extended Results}
\label{section:additional_results}
We list detailed ablations and more algorithms tested with results in \sref{Table}{table:main_results_extended}
\begin{table*}[!b]
	\small
	\centering
	\setlength\tabcolsep{0.2em}
	\begin{tabularx}{\linewidth}{lcccccc}
		\toprule
				  & \bt Compute & \bt Max Memory & \bt Params & \bt Recall@100  & \bt Recall@100   & {\bt Recall@100}  \\
		\bt Model & (vs Ours)   & (vs Ours)  & (All)      & (Production)     & (Offline)     & $(\mathrm{Gain}\%)$  \\
		\midrule
		& & &{\bt Baselines} & & &\\
		Top-N 	& 0.0x 				& 0.0x  & 0.0M / 0.0M & 0.002   & 0.0024  & ${\color{magenta}{98.74\%  \downarrow}}$ \\
		SVD 	& 2.2x 				& 0.4x  & 1.9B / 0.0M & 0.014   & 0.0151  & ${\color{magenta}{92.10\%  \downarrow}}$ \\
		NCF    	& 6.0x 				& 4.0x  & 1.9B / 2.1M & 0.021   & 0.0203  & ${\color{magenta}{89.38\%  \downarrow}}$\\
		+ Content & 7.5x    			& 5.2x  & 1.9B / 3.2M & 0.070   & 0.0626  & ${\color{red}{67.26\% \downarrow}}$ \\
		GCMC       & 15x 			& 9.0x  & 1.9B / 4.3M & -     	& 0.1229  & ${\color{red}{35.72\%  \downarrow}}$ \\
		GAT       & 24x 				& 12.0x & 1.9B / 8.2M &      	& 0.1329  & ${\color{red}{30.42\%  \downarrow}}$ \\
		+ DropoutNet       	& 26x 	& 12.1x & 1.9B / 8.2M & 0.1529  & 0.1630  & ${\color{red}{14.74\%  \downarrow}}$ \\
		LightGCN       		& 5.2x 	& 6.1x  & 1.9B / 0.0M & -     	& 0.1016  & ${\color{magenta}{46.88\%  \downarrow}}$ \\
		+ Content       		& 7.1x 	& 8.2x  & 1.9B / 1.1M & -		& 0.1912  & ${\color{black}{00.00\%  \uparrow}}$ \\
		
		\midrule
		& & & {\bt Ours} & & &\\
		
		Base        & 5.8x 			& 6.5x  & 1.9B / 0.0M & -     	& 0.1391  & ${\color{red}{27.25\%  \downarrow}}$\\
		+ Content       & 7.1x 		& 7.8x  & 1.9B / 1.1M & 0.1820  & 0.1871  & ${\color{orange}{02.14\%  \downarrow}}$ \\
		- Node embeddings   & 1.1x 	& 0.7x  & 1.1M / 1.1M & 0.1486  & 0.1426  & ${\color{red}{25.42\%  \downarrow}}$ \\
		+ Node Id hash       & 1.2x & 0.7x  & 1.1M / 1.1M & 0.1611  & 0.1591  & ${\color{orange}{16.78\%  \downarrow}}$ \\
		+ Node type hash     & 1.3x & 0.7x  & 1.1M / 1.1M & -	    & 0.1607  & ${\color{orange}{15.95\%  \downarrow}}$ \\
		+ L2-Norm WAML & 1.1x 		& 0.7x  & 1.1M / 1.1M & -     	& 0.1721  & ${\color{orange}{09.98\%  \downarrow}}$ \\
		+ tune $\alpha^{i}$ & 1.1x 	& 0.7x  & 1.1M / 1.1M & 0.2177  & 0.2211  & ${\color{yellow}{15.63\%  \uparrow}}$ \\
		+ Simple FFN & 1.4x  		& 0.8x  & 3.2M / 3.2M & -	    & 0.2574  & ${\color{lime}{34.62\%  \uparrow}}$ \\
		+ BERT-FFN & 1.8x  			& 1.0x  & 6.6M / 6.6M & -	    & 0.2720  & ${\color{green}{42.25\%  \uparrow}}$ \\
		+ L2-Norm BERT-FFN & 1.6x  	& 1.0x  & 6.6M / 6.6M & 0.2804 	& 0.2751  & ${\color{green}{43.88\%  \uparrow}}$ \\
		+ tune $\beta$ &  1.6x 		& 1.0x  & 6.6M / 6.6M & 0.2825  & 0.2835  & ${\color{green}{48.27\%  \uparrow}}$ \\
		+ Contrastive loss &   1.2x & 1.0x  & 6.6M / 6.6M & -     	& 0.3136  & ${\color{olive}{64.01\%  \uparrow}}$ \\
		+ L2-Norm output &    1.0x  & 1.0x  & 6.6M / 6.6M & 0.3116 	& 0.3179  & ${\color{olive}{66.27\%  \uparrow}}$ \\

		\midrule
		& & & {\bt Full vs \bt Filtered dataset of \sref{Sec.}{section:pg}} & & &\\
		+ WAML 			&    1.0x  	& 1.0x  & 6.6M / 6.6M & 0.3116 	& 0.3179  & ${\color{olive}{66.27\%  \uparrow}}$ \\
		+ full dataset  & 45x 		&1.0x  	& 6.6M / 6.6M &- 		& 0.4051  & \bm{${\color{olive}{111.8\%  \uparrow}}$} \\
		
		\bottomrule
	\end{tabularx}
	\caption{Results of various algorithms on our use-case.}
	\label{table:main_results_extended}
\end{table*}

\end{document}